\documentclass[11pt,a4paper]{article}
\usepackage{times,latexsym}
\usepackage{url}
\usepackage[T1]{fontenc}

\usepackage{amsmath}
\usepackage{comment}
\usepackage{makecell}
\usepackage{caption}
\usepackage{tabularx}
\usepackage{subcaption}
\usepackage{xcolor}
\usepackage{graphicx}
\usepackage{amssymb}
\usepackage{multirow}
\usepackage{url}
\usepackage{tikz}

\newcolumntype{Y}{>{\centering\let\newline\\\arraybackslash\hspace{0pt}}X}

\usepackage[acceptedWithA]{tacl2018v2} 

\usepackage{xspace,mfirstuc,tabulary}

\newif\iftaclinstructions
\taclinstructionsfalse 
\iftaclinstructions
\renewcommand{\confidential}{}
\renewcommand{\anonsubtext}{(No author info supplied here, for consistency with
TACL-submission anonymization requirements)}
\newcommand{\instr}
\fi

\iftaclpubformat 

\else

\fi

\title{Categorical Metadata Representation for Customized Text Classification}

\author{Jihyeok Kim*$^1$ \quad
  Reinald Kim Amplayo*$^2$ \\
  {\bf Kyungjae Lee$^1$ \quad
  Sua Sung$^1$ \quad
  Minji Seo$^1$ \quad
  Seung-won Hwang$^1$} \\
  \textbf{(* equal contribution)} \\
  $^1${Yonsei University} \quad $^2${University of Edinburgh} \\
  {\small \tt zizi1532@yonsei.ac.kr \quad reinald.kim@ed.ac.uk} \\
  {\small \tt \{lkj0509,dormouse,ggatalminji,seungwonh\}@yonsei.ac.kr} \\
}

\begin{document}
\maketitle
\begin{abstract}
  The performance of text classification has improved tremendously using intelligently engineered neural-based models, especially those injecting categorical metadata as additional information, e.g., using user/product information for sentiment classification. These information have been used to modify parts of the model (e.g., word embeddings, attention mechanisms) such that results can be customized according to the metadata. We observe that current representation methods for categorical metadata, which are devised for human consumption, are not as effective as claimed in popular classification methods, outperformed even by simple concatenation of categorical features in the final layer of the sentence encoder. We conjecture that categorical features are harder to 
  represent for machine use, as available context only indirectly describes the category, and even such context is often scarce (for tail category).
  To this end, we propose to use basis vectors to effectively incorporate categorical metadata on various parts of a neural-based model. This additionally decreases the number of parameters dramatically, especially when the number of categorical features is large. Extensive experiments on various datasets with different properties are performed and show that through our method, we can represent categorical metadata more effectively to customize parts of the model, including unexplored ones, and increase the performance of the model greatly.
\end{abstract}

\section{Introduction}

\begin{figure}
    \centering
    \includegraphics[width=0.47\textwidth]{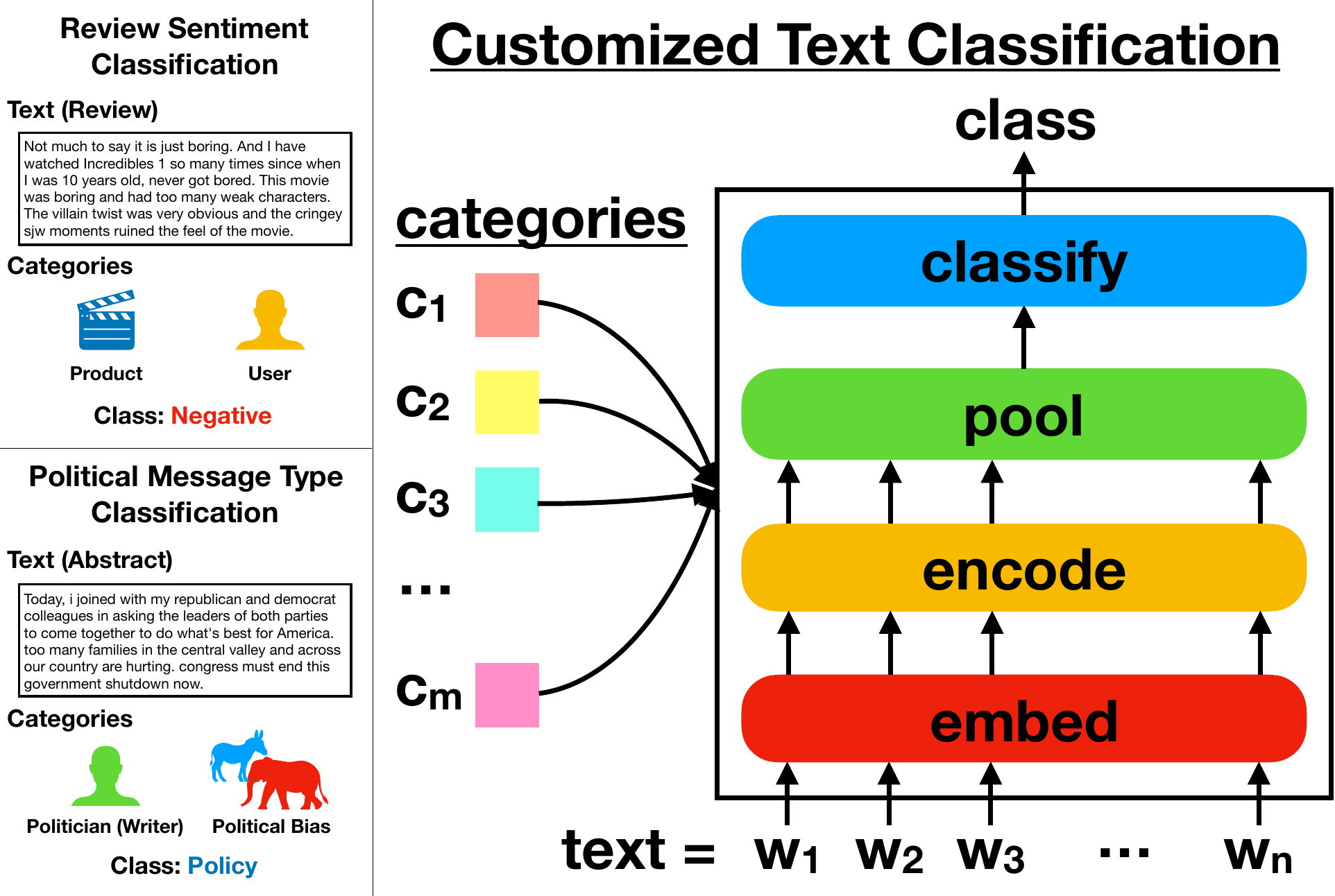}
    \caption{A high-level framework of models for the Customized Text Classification Task that inputs a text with $n$ tokens (e.g. review) and $m$ categories (e.g. users, products, etc.) and outputs a class (e.g. positive/negative). Example tasks are shown in the left of the figure.}
    \label{fig:custclass}
\end{figure}

Text classification is the backbone of most NLP tasks: review classification in sentiment analysis \cite{pang2002thumbs}, paper classification in scientific data discovery \cite{sebastiani2002machine}, and question classification in question answering \cite{li2002learning}, to name a few. While prior methods require intensive feature engineering, recent methods enjoy automatic extraction of features from text using neural-based models \cite{socher2011semi} by encoding texts into low-dimensional dense feature vectors.

This paper studies \textbf{customized text classification}, generalized from personalized text classification \cite{baruzzo2009general}, where we customize classifiers based on possibly multiple different known categorical metadata information (e.g., user/product information for sentiment classification) instead of just the user information. As shown in Figure \ref{fig:custclass}, in addition to the text, a customizable text classifier is given a list of categories specific to the text to predict its class. Existing works applied metadata information to improve the performance of a model, such as user and product \cite{tang2015learning} information in sentiment classification, and author \cite{rosen2004author} and publication \cite{joorabchi2011unsupervised} information in paper classification.

Towards the goal, we are inspired by the advancement in neural-based models, incorporating categorical information ``\textit{as is}" and injecting them on various parts of the model such as in the word embeddings \cite{tang2015learning}, attention mechanism \cite{chen2016neural,amplayo2018cold} and memory networks \cite{dou2017capturing}. Indeed, these methods theoretically make use of combined features from both textual and categorical features which make them more powerful than disconnected features. However, metadata is generated for human understanding, and thus we claim that these categories need to be carefully represented for machine use to improve the performance of the text classifier effectively.

First, we empirically invalidate the results from previous studies by showing in our experiments on multiple datasets that popular methods using metadata categories ``\textit{as is}'' perform worse than a simple concatenation of textual and categorical feature vectors. We argue that this is because of the difficulties of the model in learning optimized dense vector representation of the categorical features to be used by the classification model. The reasons are two-fold: (a) categorical features do not have direct context and thus rely solely on classification labels when training the feature vectors, and (b) there are categorical information that are sparse and thus cannot effectively learn optimal feature vectors.

Second, we suggest an alternative representation, using low-dimensional basis vectors to mitigate the optimization problems of categorical feature vectors. Basis vectors have nice properties that can solve the issues presented above because they (a) transform multiple categories into useful combinations, which serve as mutual context to all categories, and (b) intelligently initialize vectors, especially of sparse categorical information, to a suboptimal location to efficiently train them further. Furthermore, our method reduces the number of trainable parameters and thus is flexible for any kinds and any number of available categories.

We experiment on multiple classification tasks with different properties and kinds of categories available. Our experiments show that while customization methods using categorical information ``\textit{as is}'' do not perform as well as the naive concatenation method, applying our proposed basis-customization method makes them much more effective than the naive method. Our method also enables the use of categorical metadata to customize other parts of the model, such as the encoder weights, that are previously unexplored due to their high space complexity and weak performance. We show that these unexplored use of customization outperform popular and conventional methods such as attention mechanism when our proposed basis-customization method is used.

\section{Preliminaries}

\subsection{Problem: Customized text classification}

The original text classification task is defined as follows: Given a text $W=\{w_1,w_2,...,w_n\}$, we are tasked to train a mapping function $f(W)$ to predict a correct class $y\in\{y_1,y_2,...,y_p\}$ among the $p$ classes. The \textbf{customized text classification} task makes use of the categorical metadata information attached on the text to customize the mapping function. 
In this paper, we define categorical metadata as non-continuous information that describes the text\footnote{We limit our scope to texts with categorical metadata information (e.g., product reviews, news articles, tweets, etc.), which covers most of the texts in the Web. Texts without metadata can use predicted categorical information, such as topics from a topic model, which are commonly used \cite{zhao2017topic,chou2017context}. However, since the prediction may be incorrect, performance gains cannot be guaranteed. We leave the investigation of this area in future work.}.
An example task is review sentiment classification with user and product information as categorical metadata.

Formally, given a text $t=\{W,C\}$, where $W=\{w_1,w_2,...,w_n\}$, $C=\{c_{1}, c_{2}, ..., c_{m}\}$ and $w_x$ is the $x$th of the $n$ tokens in the text, and $c_{z}$ is the category label of the text on the $z$th category of the $m$ available categories, the goal of customized text classification is to optimize a function $f_{C}(W)$ to predict a label $y$, where $f_{C}(W)$ is the classifier dependent with $C$. In the example task above, $W$ is the review text, and we have $m=2$ categories where $c_{1}$ and $c_{2}$ are the user and product information.

This is an interesting problem because of the vast opportunities it brings. First, we are motivated to use categorical metadata because existing works have shown that non-textual additional information, such as POS tags \cite{go2009twitter} and latent topics \cite{zhao2017topic}, can be used as strong supplementary supervision to improve the performance of text classification. Second, while previously used additional information are found to be helpful signals, they are either domain-dependent or very noisy \cite{amplayo2018translations}. On the other hand, categorical metadata are usually factual and valid information that are either inherent (e.g., user/product information) or human-labeled (e.g., research area). Finally, the customized text classification task generalizes the personalization problem \cite{baruzzo2009general}, where instead of personalizing based on single user information, we \textit{customize} based on possibly multiple categories, which may or may not include user information. This consequently creates an opportunity to develop customizable virtual assistants \cite{papacharissi2002presentation}.

\subsection{Base classifier: \mbox{BiLSTM}}

We use a Bidirectional Long Short Term Memory (\mbox{BiLSTM}) network \cite{hochreiter1997long} as our base text classifier as it is proven to work well on classifying text sequences \cite{zhou2016text}. Although the methods that are described here apply to other effective classifiers as well, such as CNNs \cite{kim2014convolutional} and hierarchical models \cite{yang2016hierarchical}, we limit our experiments to \mbox{BiLSTM} to cover more important findings.

Our \mbox{BiLSTM} classifier starts by encoding the word embeddings using a forward and a backward \mbox{LSTM}. The resulting pairs of vectors are concatenated to get the final encoded word vectors, as shown below.
\begin{align}
    \label{eq:embedding}
    w_i &\in \mathbb{W} \\
    \label{eq:forwardlstm}
    \overrightarrow{h}_i &= LSTM_f(w_i,\overrightarrow{h}_{i-1}) \\
    \label{eq:backwardlstm}
    \overleftarrow{h}_i &= LSTM_b(w_i,\overleftarrow{h}_{i+1}) \\
    h_i &= [\overrightarrow{h}_i;\overleftarrow{h}_i]
\end{align}
Next, we pool the encoded word vectors $h_i$ into a text vector $d$ using attention mechanism \cite{bahdanau2014neural,luong2015effective}, which calculates importance scores using a latent context vector $x$ for all words, normalizes the scores using softmax, and use them to do weighted sum on encoded word vectors, as shown below.
\begin{align}
    \label{eq:attention}
    e_i &= x^\top h_i \\
    a_i &= \frac{exp(e_i)}{\sum_j exp(e_j)} \\
    d &= \sum_i h_i*a_i
\end{align}
Finally, we use a logistic regression classifier to classify labels using learned weight matrix $W^{(c)}$ and bias vector $b^{(c)}$, as shown below. 
\begin{align}
    \label{eq:classifier}
    y'=W^{(c)} d+b^{(c)}
\end{align}
We can then train our classifier using any gradient descent algorithm by minimizing the negative log likelihood of the log softmax of predicted labels $y'$ with respect to the actual labels $y$.

\subsection{Baseline 1: Concatenated \mbox{BiLSTM}}

To incorporate the categories into the classifier, a simple and naive method is to concatenate the categorical features with the text vector $d$. To do this, we create embedding spaces for the different categories and get the category vectors $c_1, c_2, ..., c_m$ based on the category labels of text $d$. We then use the concatenated vector as features for the logistic regression classifier:
\begin{align}
    \label{eq:concatlstm}
    y' = W^{(c)}[d;c_1;c_2;...;c_m] + b^{(c)}
\end{align}

\subsection{Baseline 2: Customized \mbox{BiLSTM}}

While the Concatenated \mbox{BiLSTM} easily makes use of the categories as additional features for the classifier, it is not able to leverage on the possible low-level dependencies between textual and categorical features.

There are different levels of dependencies between texts and categories. For example, when predicting the sentiment of a review ``\textit{The food is very sweet.}'' given the user who wrote the review, the classifier should give a positive label if the user likes sweet foods and a negative label otherwise. In this case, the dependency between the review and the user is on the \textit{higher level}, where we look at relationships between the full text and the categories. Another example is when predicting the acceptance of a research paper given that the research area is NLP, the classifier should focus more on NLP words (e.g., language, text) rather than less related words (e.g., biology, chemistry). In this case, the dependency between the research paper and the research area is on the \textit{lower level}, where we look at relationships between segments of text and the categories.

We present five levels of Customized \mbox{BiLSTM}, which differ on the location where we inject the categorical features, listed below from the highest level to the lowest level of dependencies between text and categories. The main idea is to impose category-specific weights, instead of a single weight at each level of the model:

\begin{enumerate}
{
\item{\textbf{Customize on the bias vector}}: At this level of customization, we look at the \textbf{general biases} the categories have towards the problem. As a concrete example, when classifying the type of message a politician wrote, he/she can be biased towards writing personal messages than policy messages. Instead of using a single bias vector $b^{(c)}$ in the logistic regression classifier (Equation \ref{eq:classifier}), we use additional multiple bias vectors for each category, as shown below. In fact, this is in spirit essentially equivalent to Concatenated BiLSTM (Equation \ref{eq:concatlstm}), where the derivation is:
\begin{align*}
    y'
    &= W_d d + b_{c_1} + ... + b_{c_m} + b^{(c)} \\
    &= W_d d + W_{c_1}c_1 + ... + W_{c_m}c_m + b^{(c)} \\
    &= W^{(c)}[d;c_1;c_2;...;c_m] + b^{(c)}
\end{align*}
\item{\textbf{Customize on the linear transformation}}: At this level of customization, we look at the \textbf{text-level semantic biases} the categories have. As a concrete example, in the sentiment classification task, the review ``\textit{The food is very sweet}'' can have a negative sentiment if the user who wrote the review does not like sweets. Instead of using a single weight matrix $W^{(c)}$ in the logistic regression classifier (Equation \ref{eq:classifier}), we use different weight matrices for each category:
\begin{equation*}
    y' = W^{(c)}_{c_1}d + W^{(c)}_{c_2}d + ... + W^{(c)}_{c_m}d + b^{(c)}
\end{equation*}
\item{\textbf{Customize on the attention pooling}}: At this level of customization, we look at the \textbf{word importance biases} the categories have. A concrete example is, when classifying a research paper, NLP words should be focus more when the research area is NLP. Instead of using a single context vector $x$ when calculating the attention scores $e$ (Equation \ref{eq:attention}), we use different context vectors for each category:
\begin{align*}
    e_i &= x_{c_1}^\top h_i + x_{c_2}^\top h_i + ... + x_{c_m}^\top h_i \\
    a &= softmax(e) \\
    d &= \sum_i h_i * a_i
\end{align*}
\item{\textbf{Customize on the encoder weights}}: At this level of customization, we look at the \textbf{word contextualization biases} the categories need. A concrete example is, given the text ``\textit{deep learning for political message classification}'', when encoding the word \textit{classification}, the BiLSTM should retain the semantics of words \textit{political message} more and forget the semantics of other words more when the research area is about politics. Instead of using a single set of input, forget, output, and memory cell weights for each LSTM (Equations \ref{eq:forwardlstm} and \ref{eq:backwardlstm}), we use multiple sets of the weights, one for each category:
\begin{equation*}
    \begin{bmatrix} 
    g_t \\ i_t \\ f_t \\ o_t
    \end{bmatrix}
    = 
    \begin{bmatrix} tanh \\ \sigma \\ \sigma \\ \sigma
    \end{bmatrix}
    \left(\sum_{0<k\leq m} W^{(e)}_{c_k} [w_t; h_{t-1}] + b\right)
\end{equation*}
\item{\textbf{Customize on the word embeddings}}: At this level of customization, we look at the \textbf{word preference biases} the categories have. For example, a user can prefer the use of word ``\textit{terribly}'' as a positive adverb rather than the more common usage of the word with negative sentiment. Instead of directly using the word vectors from the embedding space $\mathbb{W}$ (Equation \ref{eq:embedding}), we add a residual vector calculated based on a nonlinear transformation of the word vector using category-specific weights:
\begin{align}
    \label{eq:wordresidual}
    r &= tanh(W^{(w)}_{c_1} w_i + ... + W^{(w)}_{c_m} w_i) \\
    w_i &= w_i + r \nonumber
\end{align}
}
\end{enumerate}

Previous works have proposed customization on bias vectors and word embeddings \cite{tang2015learning}, and on attention pooling \cite{chen2016neural}. We are the first to introduce customization on the linear transformation matrix and the encoders. Moreover, we are the first to use residual perturbations as word meaning modification for customizing word embeddings, in which we saw better performance than using a naive affine transformation, proposed in \cite{tang2015learning}, in our prior experiments.

\section{Proposed method}

\begin{figure*}[t]
    \centering
    \includegraphics[width=0.9\textwidth]{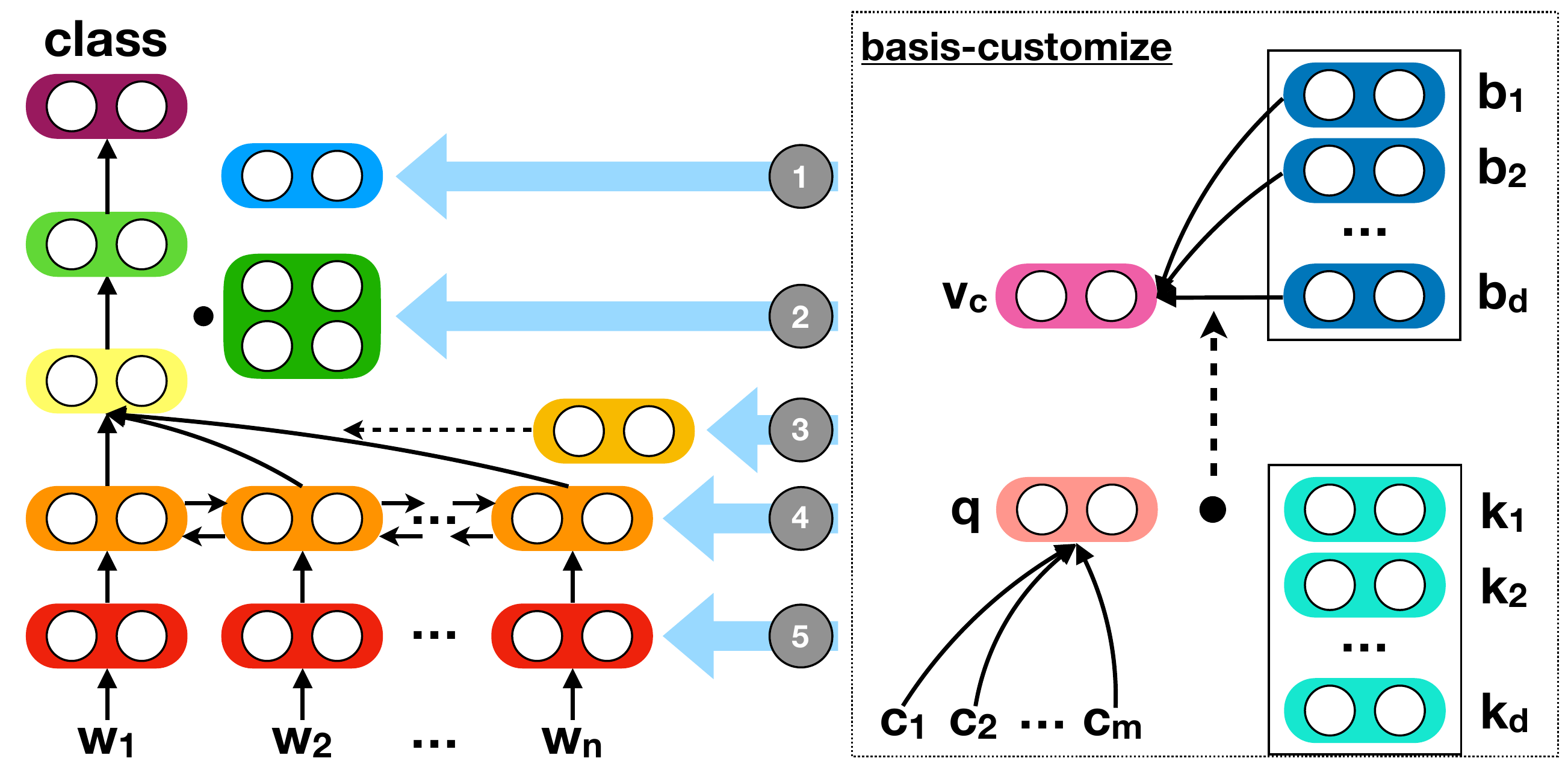}
    \caption{The full architecture of the proposed model, basis-customizing parts of the \mbox{BiLSTM} model: (1) the bias vector, (2) the linear transformation matrix, (3) the attention context vector, (4) the BiLSTM encoder weights, and (5) the word embeddings.}
    \label{fig:basismodel}
\end{figure*}

\subsection{Problems of Customized \mbox{BiLSTM}}

As explained in the previous section, Customized \mbox{BiLSTM} should perform better than Concatenated \mbox{BiLSTM}. However, that is only if the optimization of category-specific weights operates properly for machine usage. However, training the model to optimize these weights is very hard because of two reasons.

First, categorical information has unique properties that make it nontrivial to train. One property is that unlike texts which naturally use neighboring words/sentences as context \cite{lin2015hierarchical,peters2018deep}, categorical information stands alone and thus does not have information aside from itself. This forces the learning algorithm to rely solely on the classification labels $y$ to find the optimal category-specific weights. Another property is that some categories may contain labels that are sparse or do not have enough instances. For example, a user can be cold-start \cite{lam2008addressing} or does not have enough reviews. In this case, the problem expands to few-shot learning \cite{fei2006one}. Thus weights are hard to optimize using gradient-based techniques \cite{ravi2016optimization}.

Second, the number of weights is multiplied by the number of categories $m$ and the number of category labels each category has, which enlarges the number of parameters needed to be trained as $m$ increases. This magnifies the problems of context absence and information sparsity described above, since optimizing large parameters with limited inductive bias is very difficult. Moreover, because of the large parameters, some methods may not fit in commercially-available machines and thus may not be practically trainable.

\subsection{Basis Customization}

\begin{table*}[t]
  \centering
    \begin{tabularx}{\textwidth}{|l|c|l|X|}
    \hline
    Dataset & Splits & Categories & Properties \\
    \hline
    Yelp 2013 & 62,522 / 7,773 / 8,671 & \makecell[tl]{$\bullet$ users (1.6k)\\$\bullet$ products (1.6k)} & Categories can be sparse, i.e. there may not be enough reviews for each user/product).\\
    \hline
    AAPR  & 33,464 / 2,000 / 2,000 & \makecell[tl]{$\bullet$ authors (48k)\\$\bullet$ research area (144)} & Authors are sparse and have many category labels. Categories can have multiple labels (e.g. multiple authors, multidisciplinary fields). \\
    \hline
    PolMed & 4,500 / 0 / 500 & \makecell[tl]{$\bullet$ politician (505)\\$\bullet$ media source (2)\\$\bullet$ audience (2)\\$\bullet$ political bias (2)} & The dataset has more categories. Categories with binary labels may not be diverse enough to be useful. \\
    \hline
    \end{tabularx}%
  \caption{The datasets, the split sizes (train, dev, test), and the available categories and their properties. Numbers inside the parenthesis are the number of unique category labels.}
  \label{tab:dataset}%
\end{table*}%

We propose to solve all the problems above by using basis vectors to produce basis-customized weights, as shown visually in Figure \ref{fig:basismodel}. Specifically, we use a trainable set of $d \ll dim$ basis vectors $B = \{b_1, b_2, ..., b_d\}$, where $dim$ is the dimension of the original weights. Let $V_c$ be the vector search space that contains all the optimal customized weight vectors $v_c$, such that $B$ is the basis of $V_c$. Basis vectors follow the spanning property, thus we can represent all vectors in $v \in V_c$ as a linear combination of $B$, i.e. $v_c = \sum_i \gamma_i * b_i$, where the $\gamma$s are the coefficients. Moreover, since we set $d$ to a small number, we constrain the search space to a smaller vector space. Hence we can find the optimal weights in a constrained search space much faster.

To determine the $\gamma$ coefficients, we first set the concatenated category vectors of the text $q=[c_1;c_2;...;c_m]$ as the query vector, and use a trainable set of key vectors $K=\{k_1, k_2, ..., k_d\}$. We then calculate the dot product between the query and key vectors, and finally use softmax to create $\gamma$ coefficients that sum to one:
\begin{align*}
    z_i &= q^\top k_i \\
    \gamma_i &= \frac{exp(z_i)}{\sum_j exp(z_j)}
\end{align*}

We can then use the $\gamma$ coefficients to basis-customize a specific weight $v$, i.e. $v_c = \sum_i \gamma_i * b_i$. In our \mbox{BiLSTM} classifier, we can basis-customize one of the following weights: (1) the bias vector $v=b^{(c)}$ and (2) the linear transformation matrix $v=W^{(c)}$ of the logistic regression classifier in Equation \ref{eq:classifier}, (3) the context vector $v=x$ of the attention mechanism in Equation \ref{eq:attention}, (4) the BiLSTM weights $v=W^{(e)}$ in Equations \ref{eq:forwardlstm} and \ref{eq:backwardlstm}, and (5) the nonlinear transformation matrix $v=W^{(w)}$ on the residual vector in Equation \ref{eq:wordresidual} to modify the word embeddings. These correspond to the five versions of Customized BiLSTM discussed above.

Basis-customizing weights help solve the problems of customizing \mbox{BiLSTM} in three ways. First, the basis vectors serve as fuzzy clusters of all the categories, that is, we can say that two sets of category labels are similar if they have similar $\gamma$ coefficients. This information can serve as mutual context information that helps the learning algorithm find optimal weights. Second, since the search space $V_c$ is constrained, the model is forced to initialize the category vectors and look for the optimal vectors inside the constrained space. This smart initialization contributes to situate vectors of sparse categorical information to a suboptimal location and efficiently trains them further, despite the lack of instances. Finally, since we only use a very small set of basis vectors, we reduce the number of weights dramatically.

\section{Experiments}

\begin{table*}[t]
    \centering
    \begin{tabular}{|c|l|ccc|cc|cc|}
    \hline
    \multicolumn{2}{|c|}{\multirow{2}[0]{*}{Models}} & \multicolumn{3}{c|}{Yelp 2013}        & \multicolumn{2}{c|}{AAPR} & \multicolumn{2}{c|}{PolMed} \\
     \multicolumn{2}{|c|}{} & Accuracy   & RMSE & Param & Accuracy & Param & Accuracy & Param \\
    \hline
    \multicolumn{2}{|c|}{Base: BiLSTM} & 63.7  & 0.687 & 442k & 61.70 & 188k & 40.30 & 86k \\
    \hline
    \multicolumn{1}{|c|}{\multirow{2}[0]{*}{\makecell{bias vector\\(concat)}}} & cust  & 66.3  & 0.661 & 1.3m  & 65.30 & 6.3m & 40.57 & 121k \\
          & basis-cust & 66.9  & 0.654 & 653k  & 64.80 & 1.7m & 40.92 & 95k \\
    \hline
    \multicolumn{1}{|c|}{\multirow{2}[0]{*}{\makecell{linear\\trasformation*}}} & cust  & \textcolor{red}{59.6}  & \textcolor{red}{0.758} & 4.6m & 63.55 & 6.3m & \textcolor{red}{40.04} & 379k \\
          & basis-cust & \textbf{67.1}  & \textbf{0.662} & 655k  & \textbf{65.75} & 1.7m & \textbf{41.89} & 96k \\
    \hline
    \multicolumn{1}{|c|}{\multirow{2}[0]{*}{\makecell{attention\\pooling}}} & cust  & 65.4  & 0.674 & 1.3m & 62.80 & 6.3m & 40.93 & 119k \\
          & basis-cust & 66.0  & 0.671 & 652k & \textbf{65.85} & 1.7m & 41.73 & 95k \\
    \hline
    \multicolumn{1}{|c|}{\multirow{2}[0]{*}{\makecell{encoder\\weights*}}} & cust  & - & - & - & - & - & \textcolor{red}{40.26} & 43.5m \\
          & basis-cust & \textbf{66.1} & \textbf{0.665}  &  1.5m  & \textbf{66.15} & 2.1m & \textbf{41.42}  & 179k \\
    \hline
    \multicolumn{1}{|c|}{\multirow{2}[0]{*}{\makecell{word\\embedding*}}} & cust  &     \textcolor{red}{58.4} & \textcolor{red}{0.767} &  294m  & - & - & 40.84  & 46.0m \\
          & basis-cust &  \textbf{66.1} & \textbf{0.666}  &  1.0m & \textbf{65.80} & 2.0m & 41.58   & 455k \\
    \hline
    \end{tabular}%
    \caption{Accuracy, RMSE, and parameter values of competing models for all datasets. An asterisk (*) indicates customization methods first introduced in this paper. A dash (-) indicates the model is too big to be trained in an NVIDIA 1080 Ti GPU. \textbf{Bold-face} indicates that the performance of basis-customization is significantly better ($p<0.05$) than that of a simple customization. Values colored \textcolor{red}{red} are performance weaker than that of the BiLSTM model, thus customization hurts the performance in those cases.}
    \label{tab:results}%
\end{table*}

We experiment on three datasets for different tasks: (1) Yelp 2013 dataset\footnote{\url{http://ir.hit.edu.cn/~dytang}} \cite{tang2015learning} for Review Sentiment Classification, (2) AAPR dataset\footnote{\url{https://github.com/lancopku/AAPR}} \cite{yang2018automatic} for Paper Acceptance Classification, and (3) PolMed dataset\footnote{\url{https://www.figure-eight.com/}} for Political Message Type Classification. Statistics, categories, and properties of the datasets are reported in Table \ref{tab:dataset}. Details about the datasets are discussed in the next sections.

General experimental settings are as follows. The dimensions of the word vectors are set to 300. We use pre-trained GloVe embeddings \cite{pennington2014glove} to initialize our word vectors. We create UNK tokens by transforming tokens with frequency less than five into UNK. We handle unknown category labels by setting their corresponding vectors to zero. We tune the number of basis vectors $d$ using a development set, first by sweeping across 2 to 30 with large intervals, and then by searching through the neighbors of the best configuration during the first sweep. Interestingly, $d$ tends to be very small between values 2 to 4. We set the batch size to 32. We use stochastic gradient descent over shuffled mini-batches with the Adadelta update rule \cite{zeiler2012adadelta} with $l_2$ constraint of 3. We do early stopping using the accuracy of the development set. We perform 10-fold cross-validation on the training set when the development set is not available. Dataset-specific settings are described in their corresponding sections.

We compare the performance of the following competing models: the base classifier \mbox{BiLSTM} with no customization, the five versions (i.e., bias, linear, attention, encoder, embedding) of Customized \mbox{BiLSTM}, and our proposed basis-customized versions. We report the accuracy and the number of parameters of all models, and additionally report the RMSE values for the sentiment classification task. We also compare with results from previous papers whenever available. Results are shown in Table \ref{tab:results}, and further discussions are reported in the following sections.

\subsection{Review sentiment classification}

\begin{table*}[t]
  \centering
    \begin{tabular}{|c|l|cc|}
    \hline
    \multicolumn{2}{|c|}{Models} & Acc   & RMSE \\
    \hline
    UPNN \cite{tang2015learning}  & CNN + word-cust + bias-cust & 59.6  & 0.784 \\
    UPDMN \cite{dou2017capturing} & LSTM + memory-cust & 63.9  & 0.662 \\
    NSC \cite{chen2016neural}  & LSTM + attention-cust & 65.0  & 0.692 \\
    HCSC \cite{amplayo2018cold} & BiLSTM + CNN + attention-cust (CSAA) & 65.7  & \textbf{0.660} \\
    PMA  \cite{pengcheng2017parallel} & HierLSTM + attention-cust (PMA) & 65.8  & 0.668 \\
    DUPMN \cite{long2018dual} & HierLSTM + memory-cust & 66.2  & 0.667 \\
    CMA \cite{ma2017cascading} & HierAttention + attention-cust (CMA) & \textbf{66.4} & 0.677 \\
    \hline
    \multicolumn{1}{|c|}{\multirow{3}[2]{*}{Our best models}} & BiLSTM + encoder-basis-cust & 66.1  & 0.665 \\
          & BiLSTM + bias-basis-cust & 66.9  & \textbf{\underline{0.654}} \\
          & BiLSTM + linear-basis-cust & \textbf{\underline{67.1}} & 0.662 \\
    \hline
    \end{tabular}%
  \caption{Performance comparison of previous and our best models in the Yelp 2013 dataset. Our best models perform better, even though we only use a single BiLSTM encoder.}
  \label{tab:yelpresult}%
\end{table*}%

Review sentiment classification is a task of predicting the sentiment label (e.g., 1 to 5 stars) of a review text \cite{pang2002thumbs}. We use users and products as categorical metadata. One main characteristic of the categorical information here is that both user and product can be cold-start entities \cite{amplayo2018cold}. Thus issues on sparseness may aggravate. We use 256 dimensions for the hidden states in the \mbox{BiLSTM} encoder and the context vector in the attention mechanism, and 64 dimensions for each of the user and product category vectors.

The results in Table \ref{tab:results} show that when using Customized BiLSTM, customizing on the bias vector (i.e., Concatenated BiLSTM) performs the best compared to customizing on other parts of the model with lower dependencies, which is counter-intuitive and contrary to previously reported results. Moreover, the performances of customizing on the linear transformation matrix and word embedding are weaker than that of the base BiLSTM model, while customizing on the encoder weights makes the model too big to be trained in our GPU. When using our proposed basis-customization method, we obtain a significant increase in performance on all levels of customization in almost all performance metrics. Overall, a BiLSTM basis-customized on the linear transformation matrix, the bias vector, and the encoder weights perform the best among the models. Finally, we reduce the number of parameters dramatically by at least half compared to the Customized BiLSTM, which enables the training of Basis-Customized BiLSTM on encoder weights.

In addition to the competing models above, we also report results from previous state-of-the-art sentiment classification models that use user and product information:
(a) \textbf{UPNN} \cite{tang2015learning} uses a CNN encoder and customizes on bias vectors and word embeddings;
(b) \textbf{UPDMN} \cite{dou2017capturing} uses an LSTM encoder and customizes on memory vectors;
(c) \textbf{NSC} \cite{chen2016neural} uses a hierarchical LSTM encoder and customizes on attention mechanism;
(d) \textbf{HCSC} \cite{amplayo2018cold} uses a BiLSTM and a CNN as encoders and customizes on a cold-start aware attention mechanism (CSAA);
(e) \textbf{PMA} \cite{pengcheng2017parallel} uses a hierarchical LSTM encoder and customizes on PMA, an attention mechanism guided by external features;
(f) \textbf{DUPMN} \cite{long2018dual} uses a hierarchical LSTM encoder and customizes on memory vectors; and
(g) \textbf{CMA} \cite{ma2017cascading} uses a hierarchical attention-based encoder and customizes on user- and product-specific attention mechanism (CMA).
The comparison in Table \ref{tab:yelpresult} shows that our methods outperform previous models, even though (1) we only use a single BiLSTM encoder rather than more complicated ones (UPDMN and DUPMN use deep memory networks, NSC, PMA, and CMA use hierarchical encoders) and (2) we only customize on one part of the model rather than on multiple parts (UPNN customizes on bias vectors and word embeddings).

\subsection{Paper acceptance classification}

\begin{table}[t]
  \centering
    \begin{tabular}{|c|c|}
    \hline
    Models & Accuracy \\
    \hline
    \multicolumn{2}{|c|}{\textit{using full text} \cite{yang2018automatic}} \\
    \hline
    LSTM  & 60.5 \\ 
    MHCNN & \textbf{67.7} \\
    \hline
    \multicolumn{2}{|c|}{\textit{using abstract and categories (our setting)}} \\
    \hline
    LSTM  & 60.6 \\
    MHCNN & 63.7 \\
    BiLSTM & 61.7 \\
    BiLSTM+word-basis-cust & 65.8 \\
    BiLSTM+attention-basis-cust & 65.9 \\
    BiLSTM+encoder-basis-cust & \textbf{66.2} \\
    \hline
    \end{tabular}%
  \caption{Performance comparison of models using full texts and our implemented models using paper abstracts (and authors and research areas as categories for basis-customized models) as inputs in the AAPR dataset.}
  \label{tab:aaprresult}%
\end{table}%

Paper acceptance classification is a task of predicting whether the paper in question is accepted or rejected \cite{yang2018automatic}. We use the authors\footnote{In reviewing scenarios, the use of authors as additional information is discouraged for fairness. We show how powerful these features are for prediction when properly modeled, which is useful for other scenarios, e.g., deciding which arXiv papers to read.} and the research area of the papers as categorical metadata.
Both authors and research field information accept multiple labels per instance (e.g., multiple authors, multidisciplinary field), hence learning the category vector space properly is crucial to perform vector operations \cite{mikolov2013efficient}. We use 128 dimensions for both the hidden states in the \mbox{BiLSTM} encoder and the context vector in the attention mechanism and 32 dimensions for each of the categorical information. We use the paper abstract as the text. To handle multiple labels, we find that averaging the category vectors works well.

The results in Table \ref{tab:results} show similar trends from the sentiment classification results. First, we obtain better performance when using Concatenated BiLSTM compared to when using Customized BiLSTM. Second, incorporating metadata information on the attention mechanism does not perform as well as previously reported. Third, when customizing on encoder weights and word embedding, the model parameters are too big to be trained on a commercial GPU. Finally, we see significant improvements in all levels of customization when using our proposed basis-customization method, except on the bias vectors where we obtain comparable results. 
Overall, a BiLSTM basis-customized on the encoder weights, the attention pooling, and the word embedding perform the best among all the models. We also see at least 3.7x reduction of parameters when comparing Customized BiLSTM and Basis-Customized BiLSTM.

We also compare our results from previous literature \cite{yang2018automatic}, where they proposed a modular and hierarchical CNN-based encoder (MHCNN), and
used the full text (i.e., from the title and authors up to the conclusion section), instead of just the abstract, the author and the research area information. Results are reported in Table \ref{tab:aaprresult}, although full text and abstract results are not directly comparable since the original authors did not release the train/dev/test splits of their experiments.
We instead re-run MHCNN using our settings and compare with our models.
The results show that using either full text or abstract as input to LSTM produces similar results, thus using just the abstract can give us similar predictive bias when using the full text, at least in this dataset. Moreover, our best models (1) perform significantly better ($p < 0.5$) than MHCNN when restricted to our settings,
and (2) are competitive with the state-of-the-art, even though we use a simple BiLSTM encoder and only have access to the abstract, authors, and research area information.

\subsection{Political message type classification}

Political message type classification is a task of predicting the type of information a message written by a politician, is conveying, among the following nine types: attack, constituency, information, media, mobilization, personal, policy, support, and others. Two characteristics of this dataset different from others are (a) that it has four kinds of categorical information: the audience (national or constituency), bias (neutral or partisan), politician, and the source (Twitter or Facebook) information, and (b) that the category types of three categories are not diverse as they only have binary category labels. Since all of these categories may not give useful information biases to the classifier, models should be able to select which categories are informative or not. We use 64 dimensions for the hidden states in the BiLSTM encoder and the context vector in the attention mechanism, and 16 dimensions for the category vectors of each of the categorical information.

The results in Table \ref{tab:results} also show similar trends from the previous task, but since the dataset is smaller, we can compare the performance of the model when customizing on encoder weights. We show that while Customized BiLSTM on linear transformation matrix and encoder weights show weaker performance than the base BiLSTM model, Basis-Customized BiLSTM on the same levels show significantly improved performance, where Basis-Customized BiLSTM on linear transformation matrix performs the best among the competing models. The parameters also decreased dramatically, especially on encoder weights and on word embedding where we see at least 100x difference in parameter size.

\section{Analysis}

\begin{table*}[htbp]
  \small
  \centering
  \begin{subtable}{\textwidth}
    \begin{tabularx}{\textwidth}{|l|Y|Y|Y|}
    \hline
    Abstract & \multicolumn{3}{p{0.81\textwidth}|}{Several tasks in argumentation mining and debating, question-answering, and natural language inference involve classifying a sequence in the context of another sequence (referred as bi-sequence classification). For several single sequence classification tasks, the current state-of-the-art approaches are based on recurrent and convolutional neural networks. On the other hand, for bi-sequence classification problems, there is not much understanding as to the best deep learning architecture. In this paper, we attempt to get an understanding of this category of problems by extensive empirical evaluation of 19 different deep learning architectures (specifically on different ways of handling context) for various problems originating in natural language processing like debating, textual entailment and question-answering. Following the empirical evaluation, we offer our insights and conclusions regarding the architectures we have considered. We also establish the first deep learning baselines for three argumentation mining tasks.} \\
    \hline
    Research Area & cs.CL (Computation and Language) & cs.IR (Information Retrieval) & cs.CR (Cryptography and Security) \\ \hline
    Classification & Accept & Accept & Reject \\
    \hline
    \end{tabularx}%
  \end{subtable}
  ~
  \begin{subtable}{\textwidth}
    \begin{tabularx}{\textwidth}{|l|Y|Y|}
    \hline
    Message & \multicolumn{2}{p{0.81\textwidth}|}{\textless UNK\textgreater { }christmas and happy holidays from my family to yours. wishing special \textless UNK\textgreater{ } to those first responders and military personnel working to ensure our safety who are unable to be with their families this holiday season. we are all thank you for your service and dedication.} \\
    \hline
    Political Bias & Neutral & Partisan \\ \hline
    Classification & Personal & Support  \\
    \hline
    \end{tabularx}%
  \end{subtable}
  \caption{Example texts from the AAPR dataset (upper) and Political Media dataset (lower) with a variable category label (research field and political bias) that changes the classification label.
  }
  \label{tab:doclevel}%
\end{table*}%

\begin{figure}
    \centering
    \includegraphics[width=0.47\textwidth]{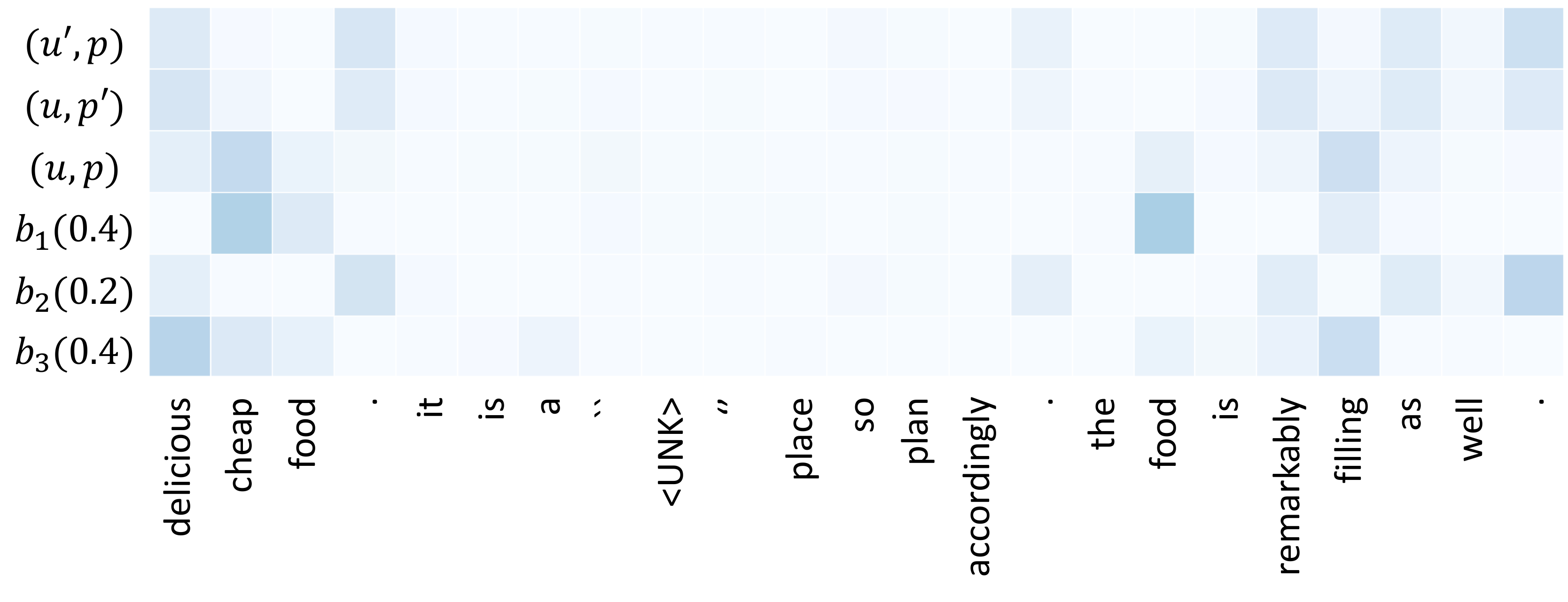}
    \includegraphics[width=0.47\textwidth]{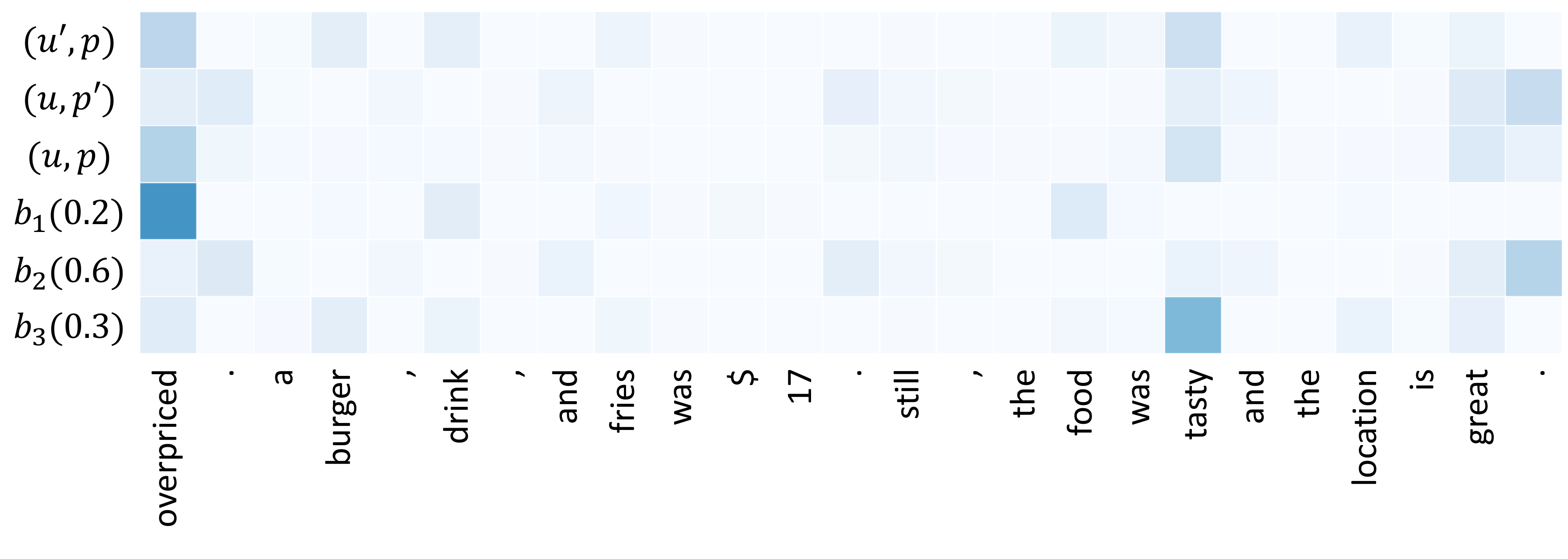}
    \caption{Examples of attention vectors from three different pairs of users and products $(u',p)$, $(u,p')$, $(u,p)$, and from the basis vectors. Numbers in parenthesis are the $\gamma_i$ coefficient of the pair $(u,p)$ with respect to basis $b_i$.}
    \label{fig:basisvecs}
\end{figure}

\begin{figure*}[t]
    \centering
    \includegraphics[width=0.24\textwidth]{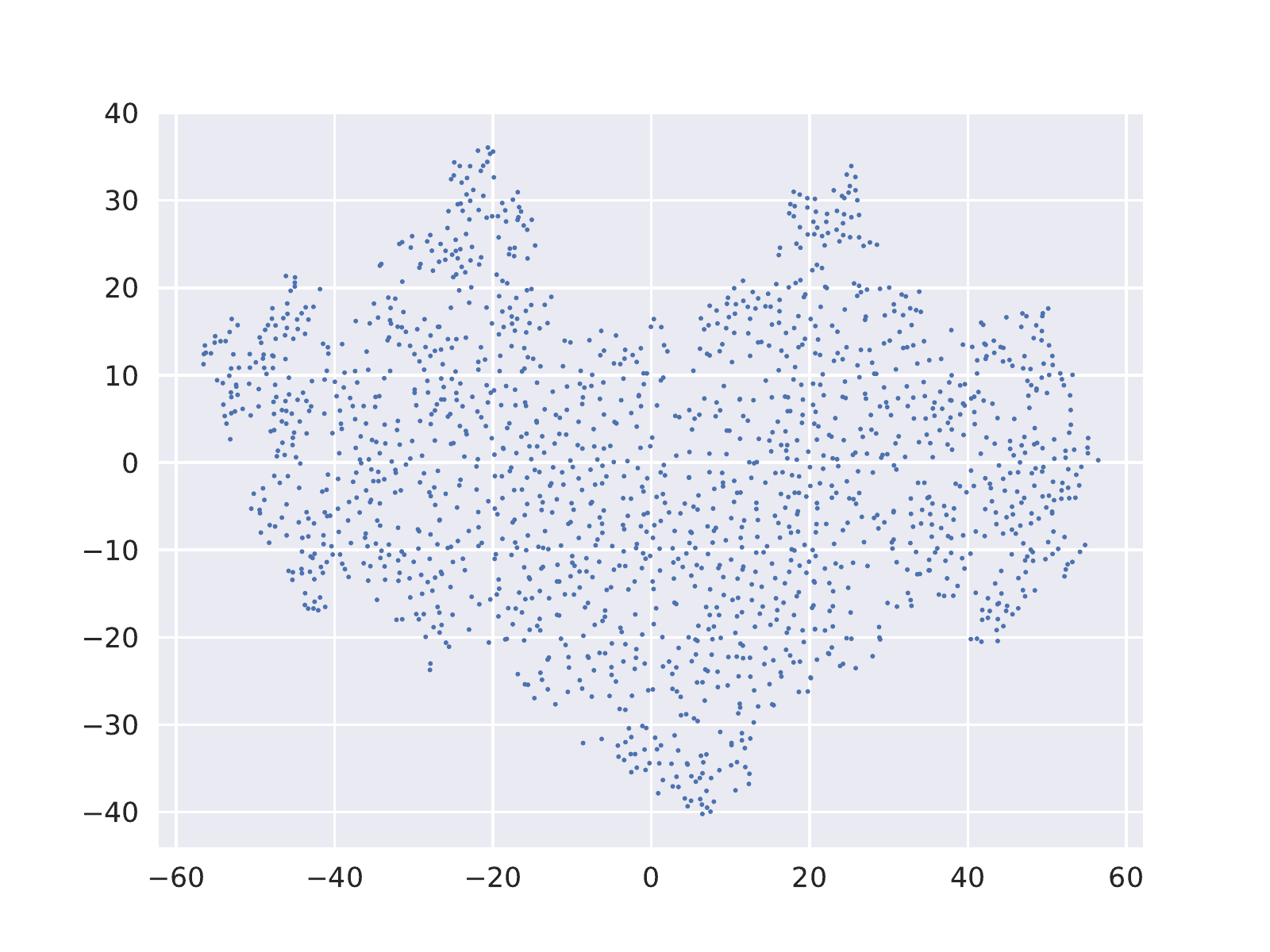}
    \includegraphics[width=0.24\textwidth]{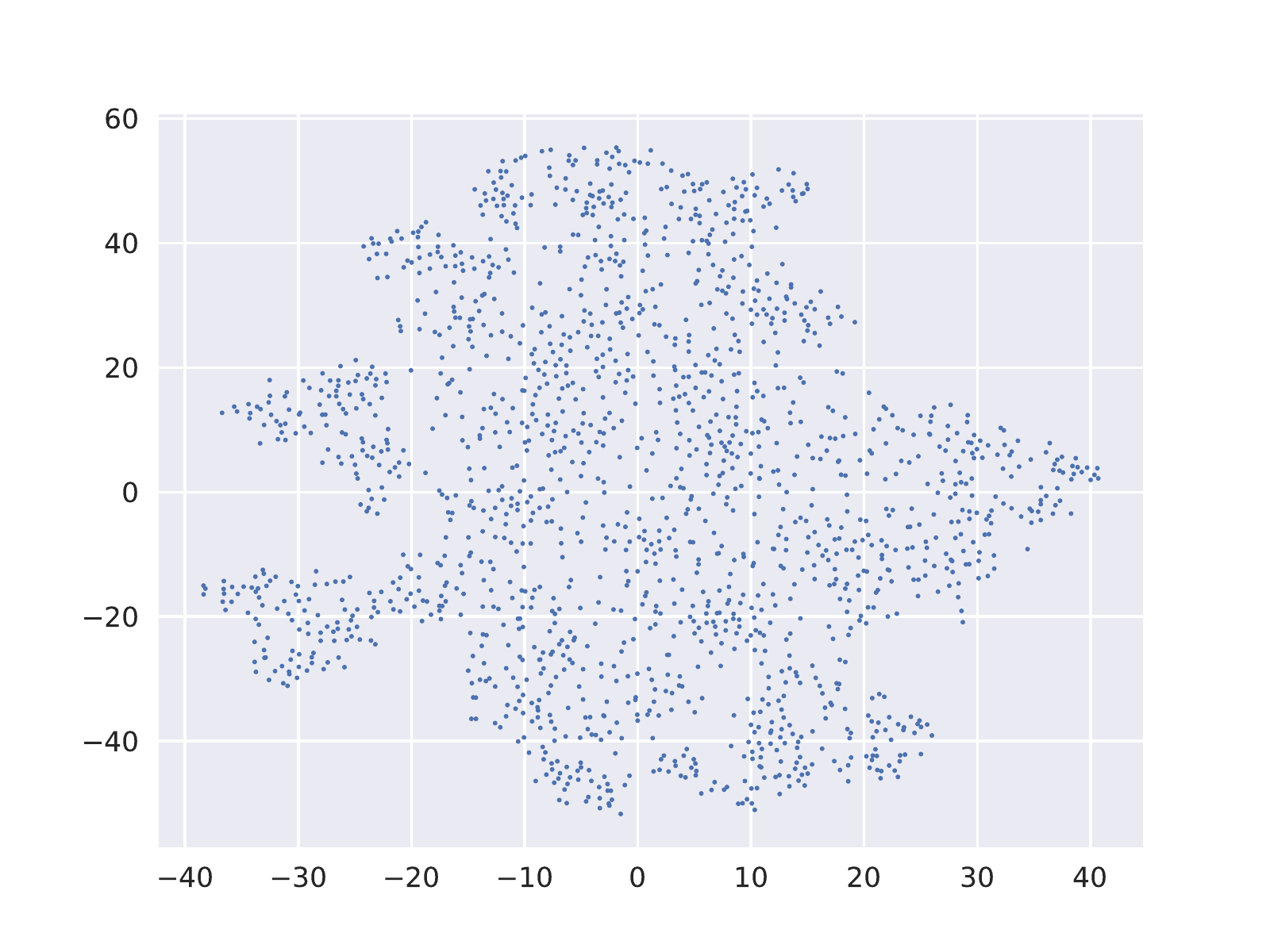}
    \includegraphics[width=0.24\textwidth]{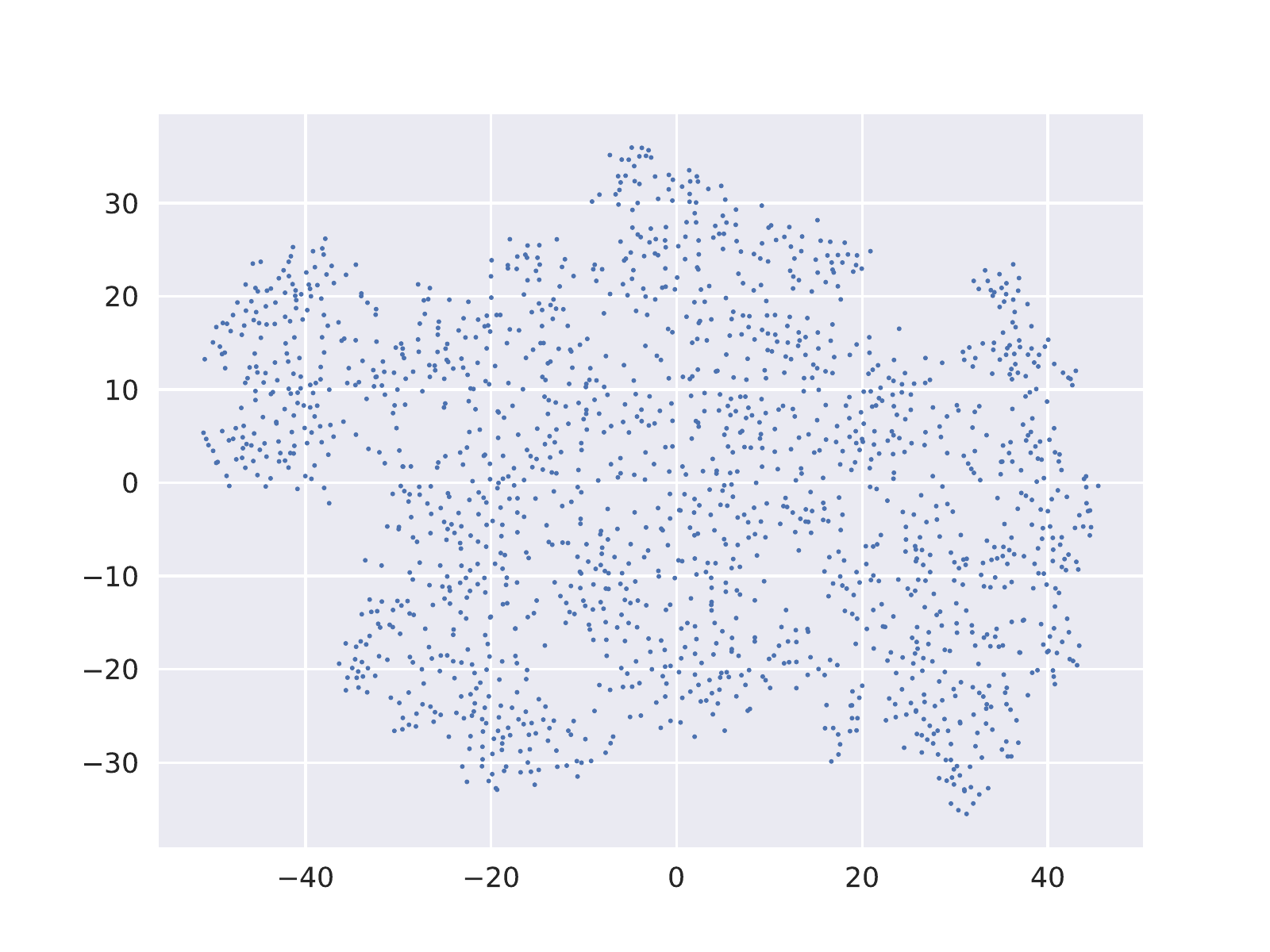}
    \includegraphics[width=0.24\textwidth]{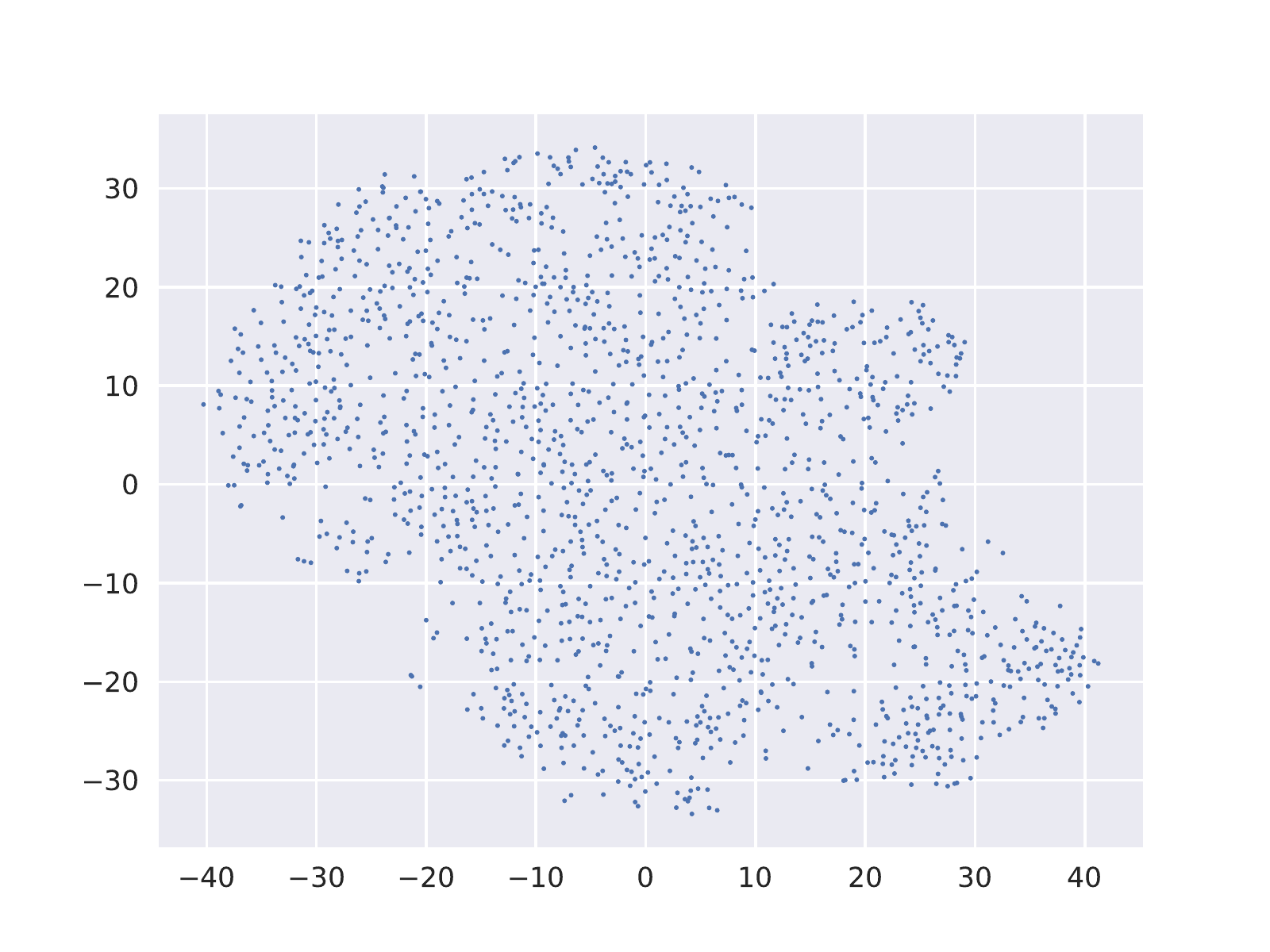}
    ~
    \includegraphics[width=0.24\textwidth]{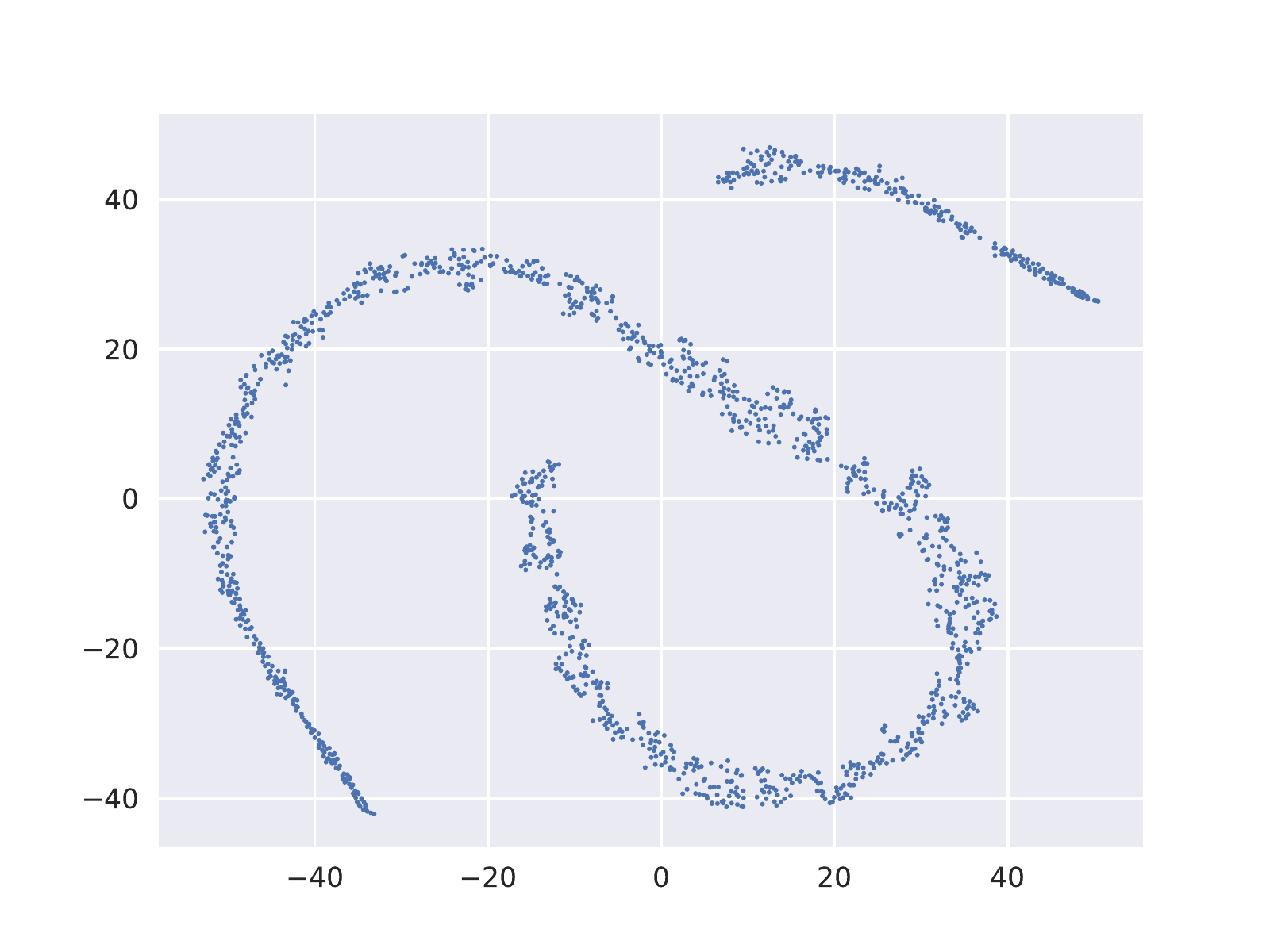}
    \includegraphics[width=0.24\textwidth]{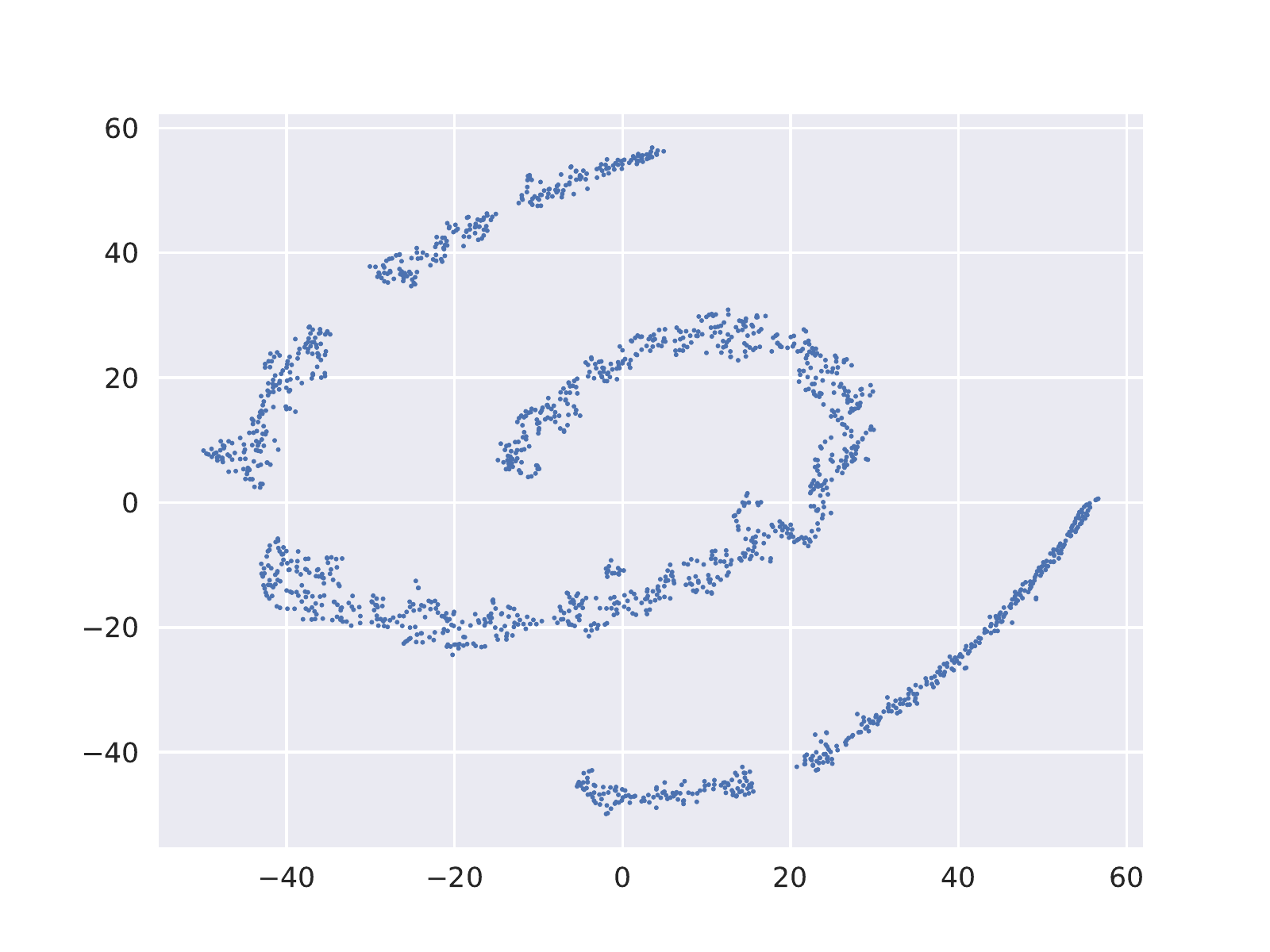}
    \includegraphics[width=0.24\textwidth]{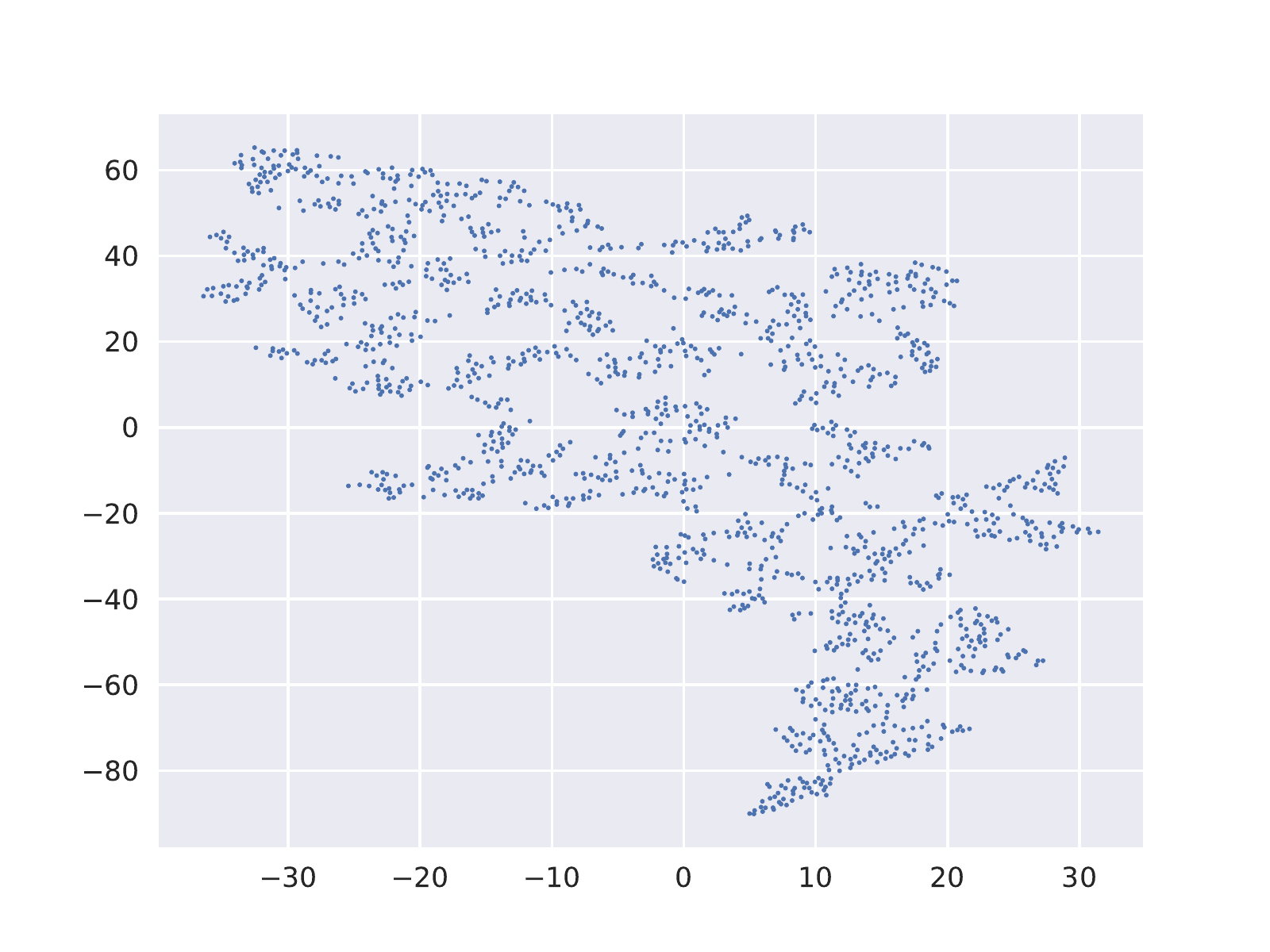}
    \includegraphics[width=0.24\textwidth]{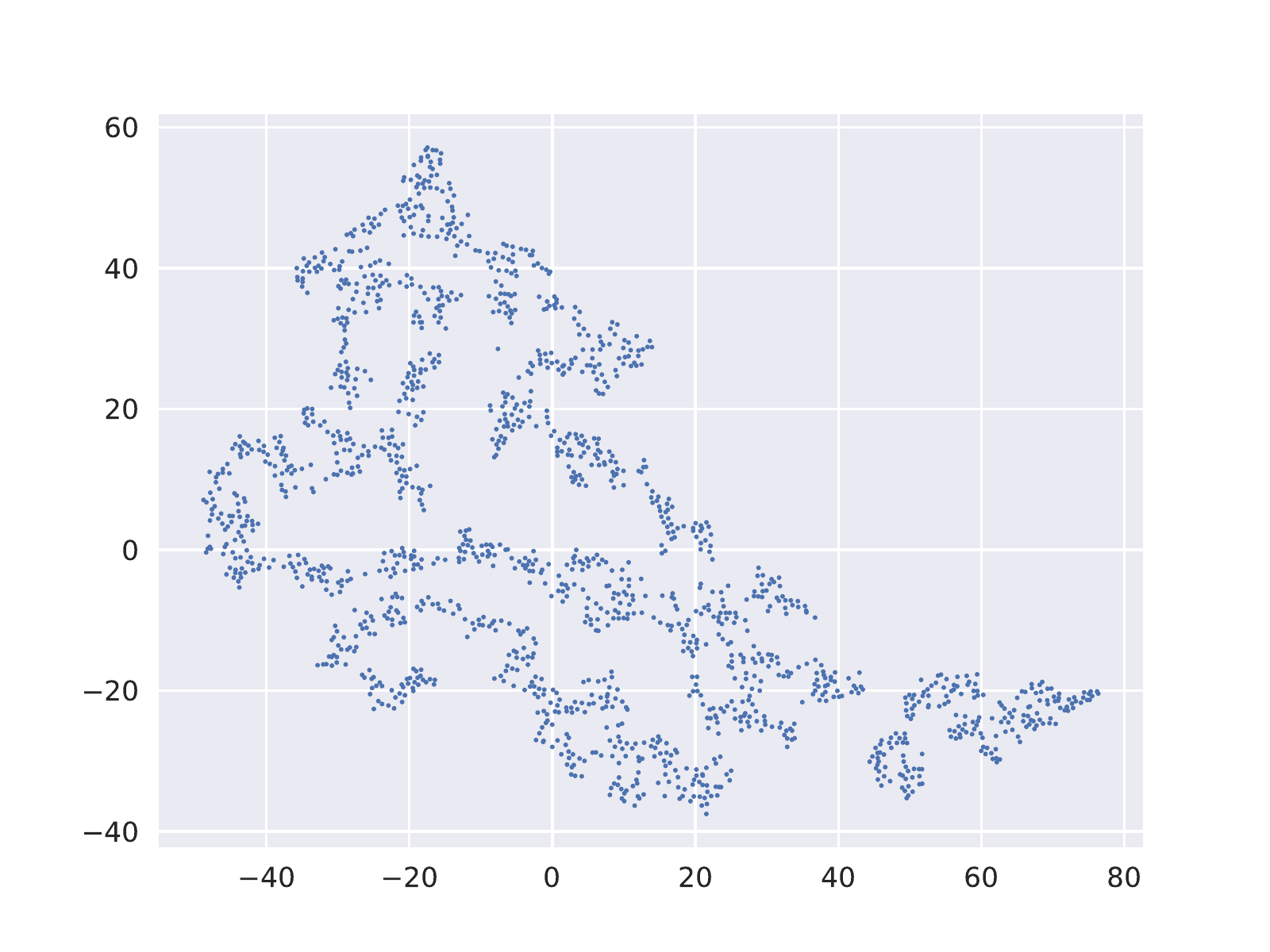}
    ~
    \includegraphics[width=0.24\textwidth]{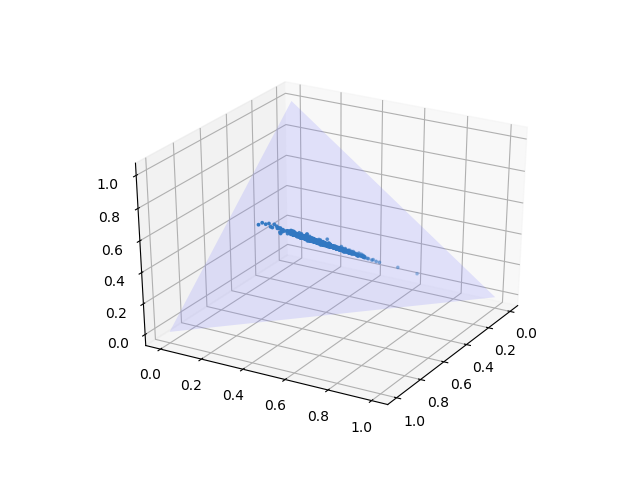}
    \includegraphics[width=0.24\textwidth]{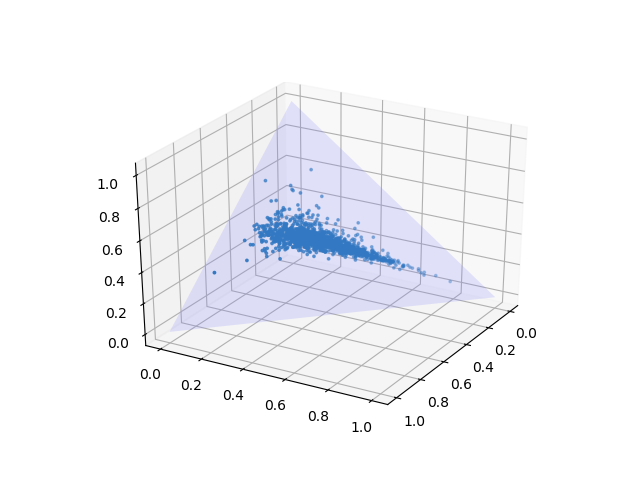}
    \includegraphics[width=0.24\textwidth]{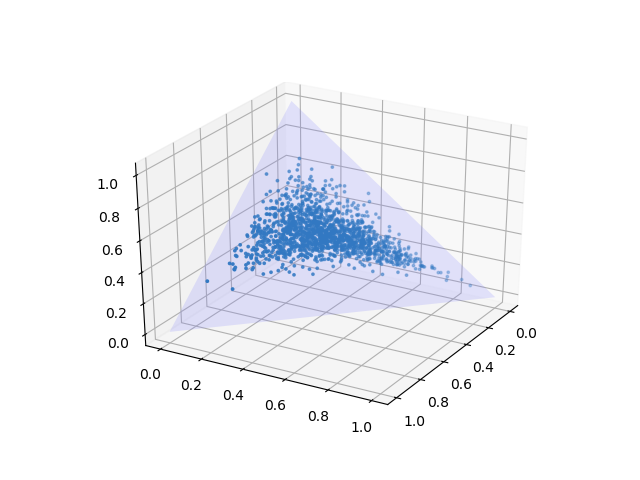}
    \includegraphics[width=0.24\textwidth]{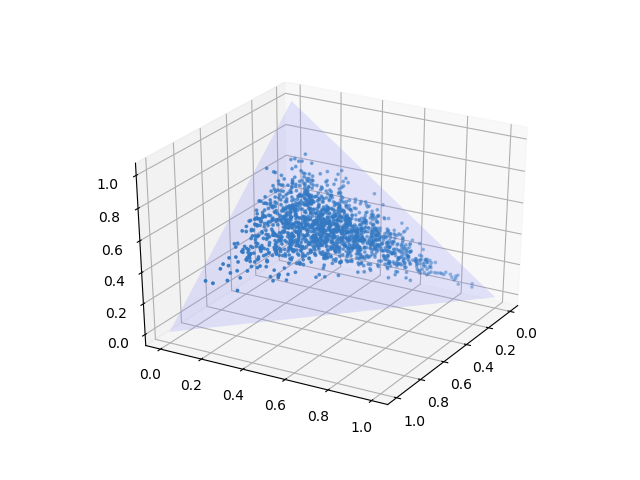}
    \caption{TSNE Visualization of the category vectors of Customized \mbox{BiLSTM} (first row) and Basis-Customized \mbox{BiLSTM} (middle row), and the $\gamma$ coefficients of the latter model (last row), when epoch is equal to 1, 2, 4, and when training has finished (left to right).
    }
    \label{fig:vectorlearn}
\end{figure*}

\subsection{Semantics of basis attention vectors}

We investigate at how basis vectors understand word-level semantics through the lens of the attention vectors they create. Previous models either combine user/product information into a single attention vector \cite{chen2016neural} or entirely separate them into distinct user and product attention vectors \cite{amplayo2018cold}. On the other hand, our model creates a single attention vector, but through the $k$ basis attention vectors, which are vectors containing fuzzy semantics among users and products. Figure \ref{fig:basisvecs} shows two examples of six attention vectors regarding a single text using the following: (1) the original user, product pair $(u,p)$, (2-3) a sampled user/product paired with the original product/user $(u',p)$ and $(u,p')$, and (4-6) the basis vectors, in the Yelp 2013 dataset. We can see in the first example that the first basis vector focuses on ``cheap'' while the third basis vector focuses on ``delicious''. An interesting output is by user $u$, such that it wants cheaper food in product $p$ yet cares more about the taste in product $p'$.

\subsection{Document-level customized dependencies}

Previous literature only focused on the analysis \cite{amplayo2018cold} and case studies \cite{chen2016neural} of word-level customized dependencies, usually through attention vectors. In this paper, we additionally investigate the document-level customized dependencies, i.e., how our basis-customization changes the document-level semantics when a category is different. Table \ref{tab:doclevel} shows two examples, one from the AAPR dataset and one from the Political Media dataset, with a variable category research area and political bias, respectively. In the first example, the abstract refers to a study on bi-sequence classification problem, a task mainly studied in the natural language processing domain, and thus gets classified as accepted when the research area category is cs.CL. The model also classifies the paper as accepted when the research area is cs.IR because the two areas are related. However, when the research area is changed to an unrelated area like cs.CR, the paper gets rejected. In the second example, the classifier predicts that when a politician with a neutral bias posts a Christmas greeting and mentions people who work on holidays, he is conveying a personal message. However, when the politician is biased towards a political party, the classifier thinks that the message is to offer support to those workers who are unable to be with their families.

\subsection{Learning strategy of basis-customized vectors}

We argue that since the basis vectors $B$ limit the search space into a constrained vector space $V_c$, then finding the optimal values of the basis-customized vectors is faster. We show in Figure \ref{fig:vectorlearn} the difference between the category vector space of Customized BiLSTM and of Basis-Customized BiLSTM. We see that the vector space of Customized BiLSTM looks random, with very few noticeable clusters, even when we iterate with four epochs. On the other hand, the basis-customized vector space starts as a cluster of one continuous spiral line, then starts to break down into smaller clusters. Multiple clusters of vectors in the vector space are clearly seen when epoch is 4. Therefore, using the basis vectors makes optimization more efficient by following the above learning strategy of starting from one cluster and dividing into smaller coherent clusters. This can also be shown in the visualization of the $\gamma$ coefficients also shown in the figure, where the coefficient values that are clumped together gradually spread out to their optimal values.

\subsection{Performance on sparse conditions}

\begin{figure*}
    \centering
    \includegraphics[width=0.47\textwidth]{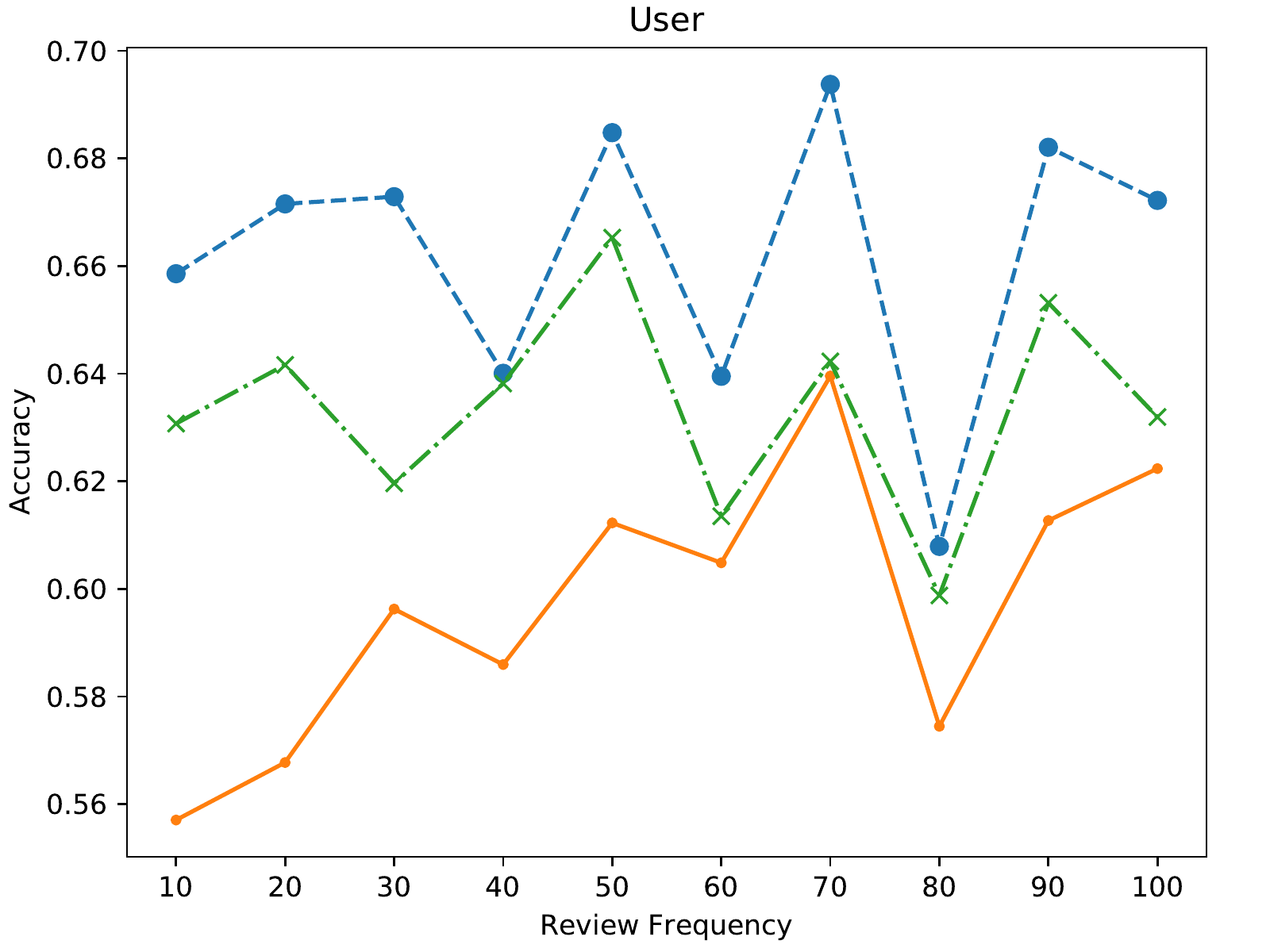}
    \includegraphics[width=0.47\textwidth]{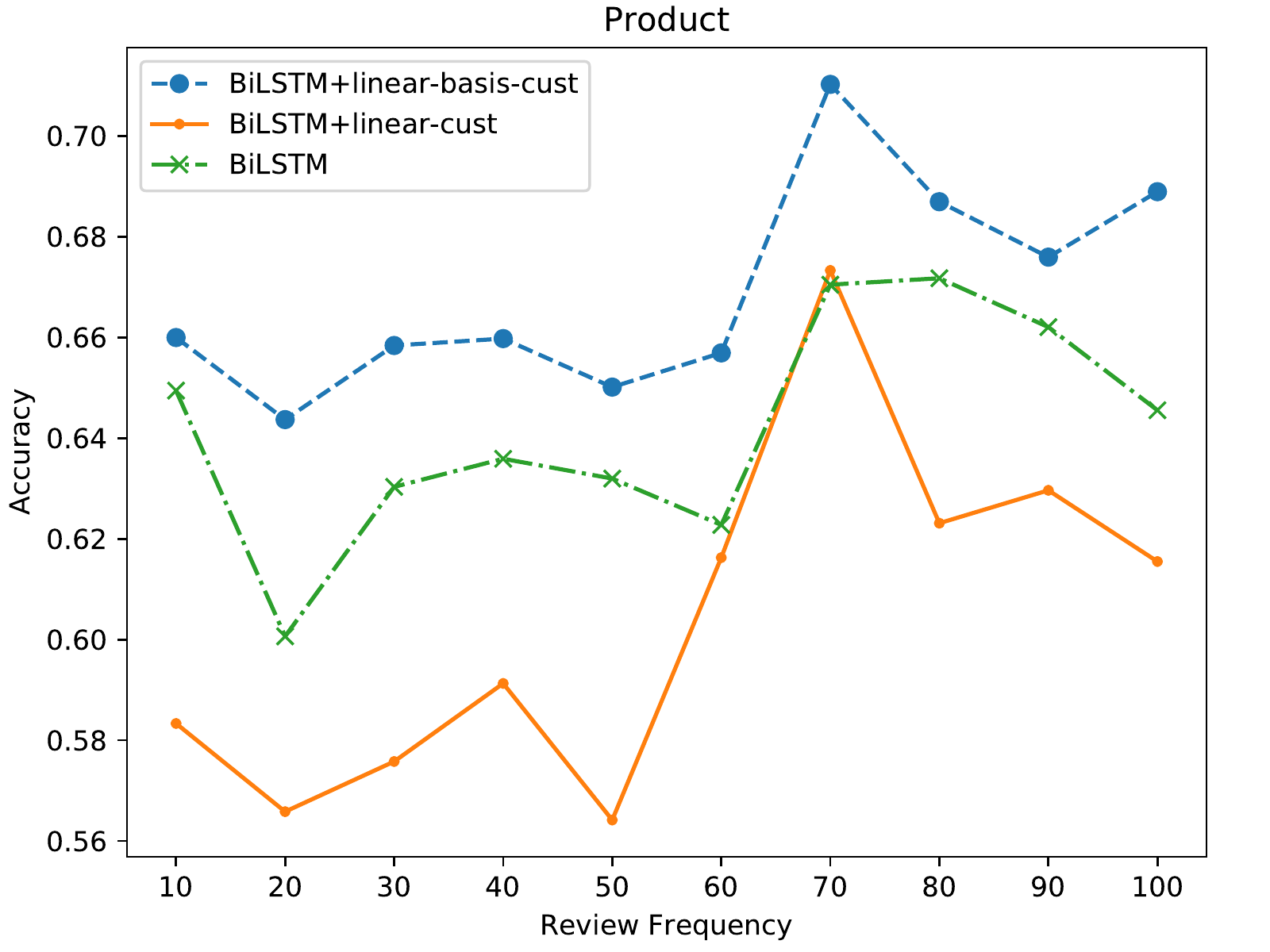}
    \caption{Accuracy per user/product review frequency on Yelp 2013 dataset. The review frequency value $f$ represents the frequencies in the range $[f, f+10)$, except when $f=100$, where it represents the frequencies in the range $[f, \inf)$.}
    \label{fig:freqperformance}
\end{figure*}

We look at the performance of three models, BiLSTM, Customized BiLSTM, and Basis-Customized BiLSTM, per review frequency of user or product. Figure \ref{fig:freqperformance} shows plots of the accuracy of the models over different user review frequency and product review frequency on the Yelp 2013 dataset. 
We observe that naive customization drops the performance of the BiLSTM model as the frequency of user/product review decreases.
This means that the model is heavily reliant on large amounts of data for optimization. On the other hand, since basis customization can learn the optimal weights of category vectors more smartly, it improves the performance of the model across all ranges of review frequency.

We finally examine the performance of our models when data contains cold-start entities (i.e., users/products may have zero or very few reviews) using the Sparse80, subset of the Yelp 2013 dataset provided in \cite{amplayo2018cold}. We compare our models with three competing models: NSC \cite{chen2016neural}, which uses a hierarchical LSTM encoder coupled with customization on the attention mechanism, BiLSTM+CSAA \cite{amplayo2018cold}, which uses a BiLSTM encoder with customization on a cold-start aware attention (CSAA) mechanism, and HCSC \cite{amplayo2018cold}, which is a combination of CNN and BiLSTM encoder with customization on CSAA. 

Results are reported in Table \ref{tab:sparse}, which provide us two observations. First, the BiLSTM model customized on the linear transformation matrix, which performs the best on the original Yelp 2013 dataset (see Table \ref{tab:yelpresult}), obtains a very sharp decrease in performance. We posit that this is because basis customization is not able to handle zero-shot cold-start entities, which are amplified in the Yelp 2013 Sparse80 dataset. 
We leave extensions of basis for zero-shot or cold-start, studied actively in machine learning~\cite{wang2019survey} and recommendation domains~\cite{sun2012survey} respectively.
Inspired by CSAA \cite{amplayo2018cold}, using similar review texts for inferring the cold-start user
(or product), we expect to infer meta context, similarly based on similar meta context, which
 may mitigate the zero-shot cold-start problem.
Second, despite having no zero-shot learning capabilities, Basis-Customized BiLSTM on the attention mechanism performs competitively with HCSC and performs better than BiLSTM+CSAA, which is Customized BiLSTM on attention mechanism with cold-start awareness.

\begin{table}[t]
    \centering
    \begin{tabular}{|c|c|}
        \hline
        Models & Accuracy \\
        \hline
        NSC & 51.1 \\
        BiLSTM+CSAA & 52.7 \\
        HCSC & \textbf{53.8} \\
        \hline
        BiLSTM+encoder-basis-cust & 50.4 \\
        BiLSTM+linear-basis-cust & 50.8 \\
        BiLSTM+bias-basis-cust & 51.9 \\
        BiLSTM+word-basis-cust & 51.9 \\
        BiLSTM+attention-basis-cust & \textbf{53.1} \\
        \hline
    \end{tabular}
    \caption{Performance comparison of competing models in the Yelp 2013 Sparse80 dataset.}
    \label{tab:sparse}
\end{table}

\section{Conclusion}

We presented a new study on customized text classification, a task where we are given, aside from the text, its categorical metadata information, to predict the label of the text, customized by the categories available. The issue at hand is that these categorical metadata information are hardly understandable and thus difficult to use by neural machines. This, therefore, makes neural-based models hard to train and optimize to find a proper categorical metadata representation. This issue is very critical, in such a way that a simple concatenation of these categorical information provides better performance than existing popular neural-based methods. We proposed to solve this problem by using basis vectors to customize parts of a classification model such as the attention mechanism and the weight matrices in the hidden layers. Our results showed that customizing the weights using the basis vectors boosts the performance of a basic BiLSTM model, and also effectively outperforms the simple yet robust concatenation methods. We share the code and datasets used in our experiments here: \url{https://github.com/zizi1532/BasisCustomize}.

\section*{Acknowledgements}
This work was supported by Microsoft Research Asia and IITP/MSIT research grant (No. 2017-0-01779).

\bibliography{tacl}
\bibliographystyle{acl_natbib}

\end{document}